\documentclass{article}
\usepackage[eatrack]{log_2022}   % for anonymous submission
\usepackage{booktabs}           % professional-quality tables
\usepackage{multirow}           % tabular cells spanning multiple rows
\usepackage{amsfonts}           % blackboard math symbols
\usepackage{graphicx}           % figures
\usepackage{duckuments}         % sample images

\usepackage[numbers,compress,sort]{natbib}

\title[]{Graph Mixer Networks}

\author[]{%
Ahmet Sarıgün\\
\institute{Middle East Technical University}\\
\email{ahmet.sarigun@metu.edu.tr}
}

\DeclareUnicodeCharacter{2212}{-}

\begin{document}

\maketitle

\begin{abstract}
In recent years, the attention mechanism has demonstrated superior performance in various tasks, leading to the emergence of GAT and Graph Transformer models that utilize this mechanism to extract relational information from graph-structured data. However, the high computational cost associated with the Transformer block, as seen in Vision Transformers, has motivated the development of alternative architectures such as MLP-Mixers, which have been shown to improve performance in image tasks while reducing the computational cost. Despite the effectiveness of Transformers in graph-based tasks, their computational efficiency remains a concern. The logic behind MLP-Mixers, which addresses this issue in image tasks, has the potential to be applied to graph-structured data as well. In this paper, we propose the Graph Mixer Network (GMN), also referred to as Graph Nasreddin Nets (GNasNets), a framework that incorporates the principles of MLP-Mixers for graph-structured data. Using a PNA model with multiple aggregators as the foundation, our proposed GMN has demonstrated improved performance compared to Graph Transformers. The source code is available publicly at \url{https://github.com/asarigun/GraphMixerNetworks}.
\end{abstract}

\section{Introduction}
Graph Neural Networks (GNNs) are a powerful tool for working with graph-structured data, which is data that is made up of entities and their relationships. GNNs have been used to solve a wide range of problems, such as node classification, link prediction, graph generation, and many others. They have attracted great interest in recent years due to their performance and the ability to extract complex information.

Graph Convolutional Networks ~\cite{kipf2016semi} (GCN) is a type of graph neural network (GNN) that uses graph convolutional layers to process data represented as graphs. GCNs can be used for various tasks such as node classification \citep{sen2008collective}, graph classification \citep{irwin2012zinc}, and link prediction. In each graph convolutional layer, the node features are updated by aggregating the features of their neighboring nodes. This is done through a convolution operation, where a linear combination of the neighboring node features is applied to each node, followed by a non-linear activation function. Graph Isomorphism Network (GIN) \citep{xu2018powerful} is another type of GNN. GIN are able to distinguish non-isomorphic graphs. GIN consists of multiple layers of neural networks, where each layer aggregates the features of the neighboring nodes using a sum pooling operation, followed by a multi-layer perceptron (MLP). GIN can be used for various tasks such as node classification, graph classification, and link prediction. Graph Attention Networks (GAT) \citep{velivckovic2017graph} is a GNN that uses attention mechanisms to assign different importance to different neighboring nodes when aggregating their features. Each node in GAT has a self-attention mechanism that allows it to weigh its own features and the features of its neighboring nodes in a learnable way. GAT can be used for various tasks such as node classification, graph classification, and link prediction. Message Passing Neural Networks (MPNN) \citep{gilmer2017neural} is a class of GNNs that generalize the idea of message passing between nodes in a graph. In MPNNs, messages are passed between nodes in the graph, and the node updates its state based on the messages received from its neighbors. MPNNs can be used for various tasks such as node classification, graph classification, and link prediction.

GCN, GIN, GAT, and MPNN are types of GNNs, each with their own characteristics and capabilities. GCN uses graph convolutional layers, GIN uses a sum pooling operation, GAT uses attention mechanisms to assign importance to different neighboring nodes, and MPNNs generalize the idea of message passing between nodes in a graph. They can all be used for various tasks such as node classification, graph classification, and link prediction.

Transformer networks have been particularly successful in natural language processing tasks such as machine translation, text summarization, and language understanding.

The key innovation of transformer networks is the use of self-attention mechanisms. In a transformer network, each element in a sequence (such as a word in a sentence) is processed by attending to all the other elements in the sequence. This allows the network to weigh the importance of different elements in the sequence when making a prediction.

Vision transformers \citep{dosovitskiy2020image} are a type of transformer network designed for computer vision tasks, adapting the transformer architecture from natural language processing to handle image data. The input image is divided into non-overlapping patches and processed individually using self-attention mechanisms. The key advantage of vision transformers is their ability to handle images of arbitrary size, unlike traditional convolutional neural networks \citep{he2016deep} which require a fixed image size.

Although transformers have shown superior performance on the image task, they have a computational quadratic complexity. MLP-Mixers \citep{tolstikhin2021mlp, melas2021you} have been used for the first time in the image task, and it has been shown to perform well without the quadratic complexity like in the attention mechanism and using only MLP. 

In this work, we propose Graph Mixer Networks (or Graph Nasreddin Networks) which uses MLP-Mixer on Graphs as a novel Graph Neural Network. They are shown to have less computational complexity than transformers and comparable performance to baseline models.

The source code is available publicly at \url{https://github.com/asarigun/GraphMixerNetworks}.

\section{Graph Mixer Networks}
\label{sec:headings}

\begin{figure}
    \centering
    \includegraphics[scale=0.40]{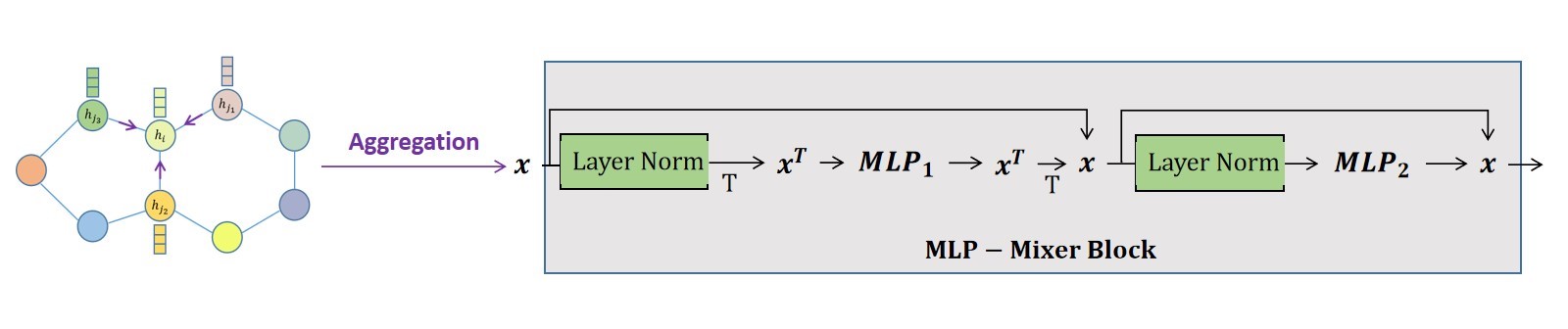}
    \caption{Mixer Layer of Graph Mixer Network where $T$ denotes transpose operator.}%%%(Figure 2)}
    \label{fig:gmn}
\end{figure}

The proposed Graph Mixer Networks (or Graph Nasreddin Networks) method leverages the increased expressivity from the multi-aggregators models such as PNA \citep{corso2020principal} and MLP-Mixer \citep{tolstikhin2021mlp, melas2021you}. GMN  architecture is given in Figure \ref{fig:gmn}.

\subsection{Motivation}

Graph Transformers use the attention mechanism. Within this mechanism, in the self-attention mechanism, the dot product between Query and Key makes the computational complexity by $O(n^2)$. For this reason, MLP-Mixers, which are applied in image tasks, are seen to be more computationally efficient and perform better than Vision Transformers \citep{tolstikhin2021mlp, melas2021you}. That is why MLP-Mixer has considered solving computational efficiency in graphs in this study. 

\subsection{Multi-aggregators}

Multiple aggregations are generalizations of sum aggregation within single aggregators and have been shown theoretically and empirically to be better at discriminating graphs. Therefore, in this study, we use multiple aggregators such as min, max, and mean aggregators used in the PNA model simultaneously.

\paragraph{Degree Scalers.} In GMN, degree scalers amplify or decrease signals based on a node's degree, allowing for more flexibility in the model. The formula for this is $S$ (scaling factor) = $\alpha$ (amplification factor) x $d$ (node degree) / $delta$ (average degree in the training set). In our research, we tested amplification factors of $-1$, $0$, and $1$, which respectively result in attenuation, no change, and signal amplification based on the node's degree.

\begin{equation}
    S(d,\alpha) = \left(\frac{\log(d+1)}{\delta}\right)^{\alpha} , d>0 , -1<\alpha<1
\end{equation}

\paragraph{Combining Aggregators.} 
Like in the PNA, we incorporate various aggregators and degree scalers using the equation $\otimes$ (Tensor product) and $\oplus$ (general aggregation function) in the GMN framework to enhance the network's flexibility.

\begin{equation}
    \oplus = \begin{pmatrix} I \\ S(D, \alpha = 1) \\ S(D, \alpha = -1) \end{pmatrix} \otimes \begin{pmatrix} Max\\ Min\\ Mean\end{pmatrix}
\end{equation}

\subsection{MLP-Mixer} 

In this work, we used the Mixer block recommended in MLP-Mixer, which is used for the image task, as the Mixer layer in GMN. Input,$x$ first goes through the layer norm operation, then transposes this matrix by the $T$ operator. The transposed $x$ passes through the Linear layer as $x^T$ and is again transposed and transformed into $x$. It is then merged with the input $x$ with the residual connection. Then it passes through Layer Norm, linear layer, and residual connection again. MLP-Mixer block can be summarized as the following equation \ref{eq:mixer}: 

\begin{equation}
\label{eq:mixer}
Mix = MLP_{2}(LayerNorm((MLP_{1}((LayerNorm(x))^{T}))^{T} + x)) + x 
\end{equation}

\subsection{Graph Mixer Network} 
As in PNA, after $h_{i}^{k}$ and $h_{j}^{k}$, which are the features of the neighboring node, are concatenated and pass through the linear layer containing learnable parameters, they pass through multiple aggregation and scalars. We take the embedding features obtained as a result of this process here as $x$. This $x$ enters the Mixer block, just like the MLP-Mixers performed in the image task. Figure 1 shows the components of the Mixer Blocks, and Equation \ref{eq:mixer} shows the Mix operator. as a result, the update function $h_{i}^{k+1}$ is obtained as in Equation \ref{eq:graphmixer}.

\begin{equation}
\label{eq:graphmixer}
h_{i}^{k+1} = Mix(h_i^k, \oplus ( h_i^k, h_j^k ,e_{ij}^k))
\end{equation}

\section{Experiments}
We evaluated the performance of GMN models on ZINC \cite{irwin2012zinc} dataset. The performance results of GMN were compared with the Message Passing Neural Networks (MPNN) ~\cite{gilmer2017neural} and Graph Transformer.  

\subsection{Dataset}
Our method was trained using the ZINC dataset, which is a dataset for predicting the solubility of chemical compounds through graph regression. The compounds in the dataset are represented as graphs, with atoms as nodes and bonds between atoms as edges. The ZINC dataset includes 12,000 molecules, with atom numbers ranging from 9 to 37. The performance of the method was evaluated using the mean absolute error (MAE) metric.

%In Appendix \ref{} we

\subsection{Results}

We trained models using the multiple aggregator(s) such as mean, max, min. Finally, we compared our results with the well-known baseline methods in the literature. The results are given in Table \ref{table:benchmarking}. Our results have shown improved performance over attention based mechanism methods.
\begin{table}[ht]
\centering
\caption{Benchmarking GMN on ZINC dataset.}
\begin{tabular}{cc} \toprule
    {\textbf{Models}} &  {\textbf{ZINC}(MAE)} \\\midrule
    
    {GCN~\cite{kipf2016semi}} & {0.367}\\
    {GAT~\cite{velivckovic2017graph}} & {0.384}\\

    {MPNN~\cite{gilmer2017neural}} & {0.288}\\
    {Graph Transformer~\cite{dwivedi2020generalization}} & {0.226}\\
    {GMN (ours)}&  {0.212}\\
    {GraphGPS~\cite{rampavsek2022recipe}}&  {0.070}\\
    \bottomrule
\label{table:benchmarking}
\end{tabular}
\end{table}
\section{Discussion and Conclusion}

The experiment shows that it is possible to create a powerful transformer-style graph regressor without using attention layers. Additionally, the MLP-Mixer model has a significant advantage over the Graph Transformer in terms of complexity \citep{tolstikhin2021mlp, melas2021you}, as it is linear in relation to the sequence length instead of quadratic. This is achieved through the use of an intermediate projection dimension within the feed-forward layer applied to aggregated learnable embeddings.

The performance of the GMN is lower than GraphGPS because the positional encoding \citep{dwivedi2021graph} and the number of parameters are quite low. In addition to its lower performance, the main disadvantage of the MLP-Mixer model is that it can only operate on sequences of a fixed length (as a result of the feed-forward layer applied to aggregated learnable embeddings). While this is not a problem in the image domain, it can be a limitation for graph neural networks because graphs do not have a fixed data structure.

This study shows that MLP-Mixers are effective for graph regression. Future research should focus on understanding the specific roles of other parts of the MLP-Mixer, such as interpretability or initialization scheme. Additionally, the report hopes to inspire further investigation into the underlying reasons for the performance of current models.

\section*{Acknowledgements}
The author express his gratitude to Ahmet S. Rifaioğlu, Gökhan Özsarı and Mehmet Volkan Atalay not just only for writing and encouragin him to write this paper but also  giving or the valuable insights and discussions.

%\newpage
\bibliographystyle{unsrtnat}
\bibliography{reference}

\begin{thebibliography}{14}
\providecommand{\natexlab}[1]{#1}
\providecommand{\url}[1]{\texttt{#1}}
\expandafter\ifx\csname urlstyle\endcsname\relax
  \providecommand{\doi}[1]{doi: #1}\else
  \providecommand{\doi}{doi: \begingroup \urlstyle{rm}\Url}\fi

\bibitem[Kipf and Welling(2016)]{kipf2016semi}
Thomas~N Kipf and Max Welling.
\newblock Semi-supervised classification with graph convolutional networks.
\newblock \emph{arXiv preprint arXiv:1609.02907}, 2016.

\bibitem[Sen et~al.(2008)Sen, Namata, Bilgic, Getoor, Galligher, and
  Eliassi-Rad]{sen2008collective}
Prithviraj Sen, Galileo Namata, Mustafa Bilgic, Lise Getoor, Brian Galligher,
  and Tina Eliassi-Rad.
\newblock Collective classification in network data.
\newblock \emph{AI magazine}, 29\penalty0 (3):\penalty0 93--93, 2008.

\bibitem[Irwin et~al.(2012)Irwin, Sterling, Mysinger, Bolstad, and
  Coleman]{irwin2012zinc}
John~J Irwin, Teague Sterling, Michael~M Mysinger, Erin~S Bolstad, and Ryan~G
  Coleman.
\newblock Zinc: a free tool to discover chemistry for biology.
\newblock \emph{Journal of chemical information and modeling}, 52\penalty0
  (7):\penalty0 1757--1768, 2012.

\bibitem[Xu et~al.(2018)Xu, Hu, Leskovec, and Jegelka]{xu2018powerful}
Keyulu Xu, Weihua Hu, Jure Leskovec, and Stefanie Jegelka.
\newblock How powerful are graph neural networks?
\newblock \emph{arXiv preprint arXiv:1810.00826}, 2018.

\bibitem[Veli{\v{c}}kovi{\'c} et~al.(2017)Veli{\v{c}}kovi{\'c}, Cucurull,
  Casanova, Romero, Lio, and Bengio]{velivckovic2017graph}
Petar Veli{\v{c}}kovi{\'c}, Guillem Cucurull, Arantxa Casanova, Adriana Romero,
  Pietro Lio, and Yoshua Bengio.
\newblock Graph attention networks.
\newblock \emph{arXiv preprint arXiv:1710.10903}, 2017.

\bibitem[Gilmer et~al.(2017)Gilmer, Schoenholz, Riley, Vinyals, and
  Dahl]{gilmer2017neural}
Justin Gilmer, Samuel~S Schoenholz, Patrick~F Riley, Oriol Vinyals, and
  George~E Dahl.
\newblock Neural message passing for quantum chemistry.
\newblock In \emph{International conference on machine learning}, pages
  1263--1272. PMLR, 2017.

\bibitem[Dosovitskiy et~al.(2020)Dosovitskiy, Beyer, Kolesnikov, Weissenborn,
  Zhai, Unterthiner, Dehghani, Minderer, Heigold, Gelly,
  et~al.]{dosovitskiy2020image}
Alexey Dosovitskiy, Lucas Beyer, Alexander Kolesnikov, Dirk Weissenborn,
  Xiaohua Zhai, Thomas Unterthiner, Mostafa Dehghani, Matthias Minderer, Georg
  Heigold, Sylvain Gelly, et~al.
\newblock An image is worth 16x16 words: Transformers for image recognition at
  scale.
\newblock \emph{arXiv preprint arXiv:2010.11929}, 2020.

\bibitem[He et~al.(2016)He, Zhang, Ren, and Sun]{he2016deep}
Kaiming He, Xiangyu Zhang, Shaoqing Ren, and Jian Sun.
\newblock Deep residual learning for image recognition.
\newblock In \emph{Proceedings of the IEEE conference on computer vision and
  pattern recognition}, pages 770--778, 2016.

\bibitem[Tolstikhin et~al.(2021)Tolstikhin, Houlsby, Kolesnikov, Beyer, Zhai,
  Unterthiner, Yung, Steiner, Keysers, Uszkoreit, et~al.]{tolstikhin2021mlp}
Ilya~O Tolstikhin, Neil Houlsby, Alexander Kolesnikov, Lucas Beyer, Xiaohua
  Zhai, Thomas Unterthiner, Jessica Yung, Andreas Steiner, Daniel Keysers,
  Jakob Uszkoreit, et~al.
\newblock Mlp-mixer: An all-mlp architecture for vision.
\newblock \emph{Advances in Neural Information Processing Systems},
  34:\penalty0 24261--24272, 2021.

\bibitem[Melas-Kyriazi(2021)]{melas2021you}
Luke Melas-Kyriazi.
\newblock Do you even need attention? a stack of feed-forward layers does
  surprisingly well on imagenet.
\newblock \emph{arXiv preprint arXiv:2105.02723}, 2021.

\bibitem[Corso et~al.(2020)Corso, Cavalleri, Beaini, Li{\`o}, and
  Veli{\v{c}}kovi{\'c}]{corso2020principal}
Gabriele Corso, Luca Cavalleri, Dominique Beaini, Pietro Li{\`o}, and Petar
  Veli{\v{c}}kovi{\'c}.
\newblock Principal neighbourhood aggregation for graph nets.
\newblock \emph{arXiv preprint arXiv:2004.05718}, 2020.

\bibitem[Dwivedi and Bresson(2020)]{dwivedi2020generalization}
Vijay~Prakash Dwivedi and Xavier Bresson.
\newblock A generalization of transformer networks to graphs.
\newblock \emph{arXiv preprint arXiv:2012.09699}, 2020.

\bibitem[Ramp{\'a}{\v{s}}ek et~al.(2022)Ramp{\'a}{\v{s}}ek, Galkin, Dwivedi,
  Luu, Wolf, and Beaini]{rampavsek2022recipe}
Ladislav Ramp{\'a}{\v{s}}ek, Mikhail Galkin, Vijay~Prakash Dwivedi, Anh~Tuan
  Luu, Guy Wolf, and Dominique Beaini.
\newblock Recipe for a general, powerful, scalable graph transformer.
\newblock \emph{arXiv preprint arXiv:2205.12454}, 2022.

\bibitem[Dwivedi et~al.(2021)Dwivedi, Luu, Laurent, Bengio, and
  Bresson]{dwivedi2021graph}
Vijay~Prakash Dwivedi, Anh~Tuan Luu, Thomas Laurent, Yoshua Bengio, and Xavier
  Bresson.
\newblock Graph neural networks with learnable structural and positional
  representations.
\newblock \emph{arXiv preprint arXiv:2110.07875}, 2021.

\end{thebibliography}
% For bibLaTeX users:
% \printbibliography
\newpage
\appendix

%\section{MMA Model Details}
%\label{subsec:Model}
%In Equation \eqref{eq:mask}, $m_{ij}^{l+1}$ represents the learned mask and $l$ represents the $l$th layer. Let $h_{i}^{l}$, $h_{j}^{l}$ and $e_{ij}^{l}$ be in $\mathbb{R}^{N}$, and then the concatenation of these vectors are in $\mathbb{R}^{3XN}$. $W_m$ is represented in $\mathbb{R}^{TX3}$. The multiplication of the concatenated $h_{i}^{l}$, $h_{j}^{l}$ and $e_{ij}^{l}$ with $W_m$ results in $\mathbb{R}^{TXN}$ dimension which gives the final dimension of $m_{ij}^{l+1}$. $T$ represents the number of hidden units.

\end{document}